\documentclass[letterpaper, 10 pt, conference]{ieeeconf}

\usepackage{amsmath}
\usepackage{amssymb}

\usepackage{tabularx}
\usepackage{graphicx}
\usepackage{bm}
\usepackage{color}
\usepackage{float}
\usepackage{units}
\usepackage{url}
\usepackage{hyperref}
\usepackage{balance}
\usepackage{amsmath}

\makeatletter 
\def\endfigure{\end@float} 
\def\endtable{\end@float}
\makeatother

\newcommand{\pos}{\bm{p}}

\usepackage{subcaption}
\usepackage{caption}

\renewcommand{\unit}[1]{{\rm #1} }

\newcommand{\rev}[1]{\textcolor{black}{#1}}

\newcommand{\gpt}[1]{\textcolor{black}{#1}}

\newtheorem{remark}{\textbf{Remark}}
\IEEEoverridecommandlockouts

\begin{document} 

\title{\Large \bf 			
Multi-contact MPC for Dynamic Loco-manipulation on Humanoid Robots
}



\author{Junheng Li and Quan Nguyen\thanks{Junheng Li and Quan Nguyen are with the Department of Aerospace and Mechanical Engineering, University of Southern California, Los Angeles, CA 90089.
email:{\tt\small junhengl@usc.edu, quann@usc.edu}}%
}%
	
\maketitle

\begin{abstract}

This paper presents a novel method to control humanoid robot dynamic loco-manipulation with multiple contact modes via multi-contact Model Predictive Control (MPC) framework.
\gpt{The proposed framework includes a multi-contact dynamics model capable of capturing various contact modes in loco-manipulation, such as hand-object contact and foot-ground contacts. Our proposed dynamics model represents the object dynamics as an external force acting on the system, which simplifies the model and makes it feasible for solving the MPC problem.}
In numerical validations, our multi-contact MPC framework only needs contact timings of each task and desired states to give MPC the knowledge of changes in contact modes in the prediction horizons in loco-manipulation. 
The proposed framework can control the humanoid robot to complete multi-tasks dynamic loco-manipulation applications such as efficiently picking up and dropping off objects while turning and walking.

\end{abstract}


\section{Introduction}
\label{sec:Introduction}

Object manipulation control on intelligent robotic systems can benefit many industries pertaining to logistics, sorting, and warehousing. There have been many successful manipulation-oriented legged robotic platforms in recent years (e.g., Digit humanoid robot \cite{digityoutube}, ANYmal quadruped with arm \cite{ferrolho2022roloma}\cite{chiu2022collision},  Spot mini with arm \cite{zimmermann2021go}, and Handle \cite{handleyoutube}). 
\gpt{The study of loco-manipulation on legged robots is fascinating due to the maneuverability and large range of motion that these robots have. This opens up numerous possibilities for implementing diverse control strategies and robot configurations to manipulate objects. For instance, there are examples of humanoid robots pushing objects with their hands \cite{murooka2021humanoid}, as well as quadruped robots using both their arms and legs to manipulate objects \cite{wolfslag2020optimisation}.}

We are particularly interested in the problem of humanoid robots carrying and manipulating an object during locomotion (i.e. applications in logistics). Agility Robotics \cite{digityoutube} and Boston Dynamics \cite{handleyoutube} allowed their humanoid robots to load and unload objects while standing still. 
\gpt{By incorporating contact schedules and ensuring smooth transitions between contact modes in dynamics models, we can explore the boundaries of humanoid loco-manipulation through the use of MPC. This includes tasks such as carrying heavy loads and throwing objects aggressively, all while maintaining stable locomotion through a Multi-contact MPC framework.}

\begin{figure}[t]
		\center
		\includegraphics[width=0.65 \columnwidth]{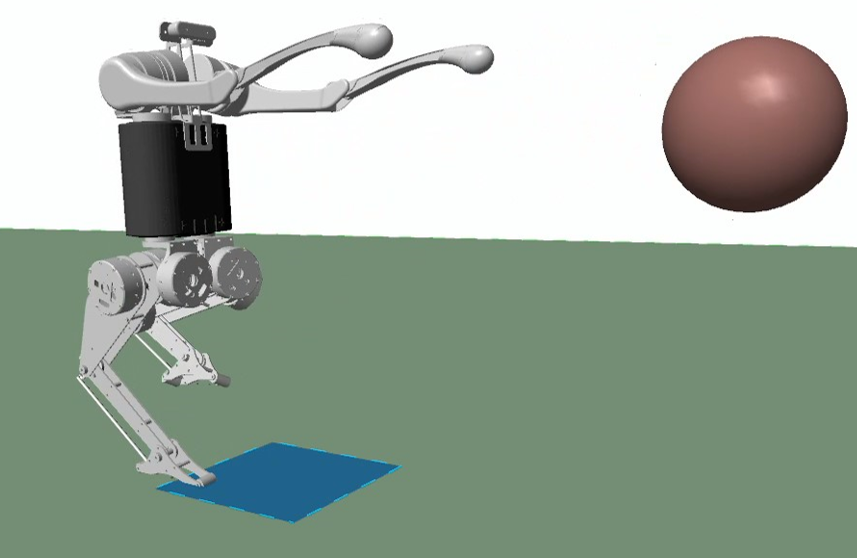}
		\caption{{\bfseries Humanoid Robot Throwing a 2 $\unit{kg}$ Ball while Walking in Place }  Simulation video: \protect\url{https://youtu.be/V8PIpE2YGhw}. }
		\label{fig:title}
		\vspace{-1.5em}
\end{figure}

Our recent work \cite{li2021force} on force-and-moment-based MPC schemes employs a simplified rigid body dynamics model and has allowed \rev{a mini bipedal robot with 5-Degree-of-Freedom (DoF) legs} to perform stable 3-D locomotion. However, simply extending the control scheme from the previous work does not work well on our humanoid robot in loco-manipulation tasks. In this paper, we develop a new humanoid robot dynamics model to consider different contact modes and use MPC as a solution to bridge the transition between different contact modes. 
\gpt{To do so, we employ contact schedules to represent contact modes in robot dynamics, and incorporate mode transitions within the MPC prediction horizons. Our humanoid loco-manipulation involves various contact modes, including hand-object contact, foot-ground contact, object-ground contact, and any combination of these modes. We use a simplified rigid body dynamics (SRBD) model that simplifies the object dynamics to minimize the need for significant changes in the MPC formulation during transitions between contact modes, while maintaining high performance in loco-manipulation.}

\rev{MPC framework has been successfully implemented on many modern legged robots for dynamic motions \cite{katz2019mini,kim2019highly,chignoli2021humanoid,li2023dynamic}. }
In \cite{chignoli2021humanoid}, the humanoid robot is able to perform aerial motions such as 3-D jumps and flips with offline kino-dynamics-based optimization and uses the MPC and Whole-body Control (WBC) as landing control. \rev{In our previous work \cite{li2023dynamic}, we also demonstrated dynamic locomotion over stepping stones on bipedal robots with a force-and-moment-based MPC control scheme.} In this work, we decide to use MPC as the main control scheme and develop it further to consider and bridge contact mode transitions and changes in humanoid loco-manipulation, specifically for the purpose of enhancing the efficiency and mobility in these tasks. 

\rev{Control strategies for multi-contact motion on humanoid robot have been investigated in many studies \cite{audren2014model,marcucci2017approximate,henze2014posture}. Authors in \cite{audren2014model} utilized a multi-contact stance planner prior to MPC to compute a series of feasible stance for bridging multi-contact transitions. 
In \cite{marcucci2017approximate}, a piecewise affine model is applied in hybrid MPC to achieve balancing  on humanoid robots that have multiple contact locations with the environment.
In our work, we focus specifically on controlling humanoid robots to carry and manipulate objects that may have multiple contact modes to the ground and to the robot during motion. In carrying and lifting scenarios, we choose to combine the object and the robot upper body in dynamics model. This approach minimizes the number of control inputs in MPC and reduces the complexity of the hybrid dynamics system to ensure problem simplicity, linearity, and feasibility. }


The main contributions of the paper are as follows:
\begin{itemize}
    \item We investigate and compare humanoid SRBD models to include object dynamics and time-varying contact modes in multi-contact MPC. 
    
    \item We propose a multi-contact MPC framework for humanoid robots to perform loco-manipulation tasks such as transporting and throwing weighted objects with contact schedule information.
    
    \item \gpt{Through numerical simulations, we have validated our proposed framework and demonstrated its ability to perform multi-task loco-manipulation on humanoid robots. This includes tasks such as throwing weighted objects while walking and maintaining balance, transporting packages more efficiently with less body yaw, and carrying heavy loads. }

\end{itemize}

The rest of the paper is organized as follows. Section. \ref{sec:robotModel} introduces the physical design and parameters of the humanoid robot and the overview of the system architecture. Section.\ref{sec:dynamicsModel} presents the dynamics models we investigated for our control framework and the multi-contact MPC in detail. Some simulation result highlights and comparisons are presented in Section. \ref{sec:Results}.

\section{Robot Model and System Overview}
\subsection{Robot Model}
\label{sec:robotModel}

In this section, we present the humanoid robot model that is used for this work. Our humanoid robot model, shown in Figure. \ref{fig:title}, is modified from the design in our previous work \cite{li2021force}, a small-scale humanoid robot with \rev{5-DoF legs and 3-DoF arms}. 
Each joint is actuated by Unitree A1 torque-controlled motor which has a 33.5 $\unit{Nm}$ maximum torque output and 21.0 $\unit{rad/s}$ maximum joint speed output. Due to the intended application, we halved the gear ratio for knee motors to have 67 $\unit{Nm}$ maximum torque.


\gpt{Our humanoid limb design strategically places joint actuators close to the trunk for mass concentration and to minimize limb dynamics during locomotion. Negligible limb mass is crucial to our force-and-moment-based simplified dynamics model used in MPC \cite{li2021force}. The humanoid robot has a total mass of 17 $\unit{kg}$.}

\subsection{System Overview}
\label{sec:sysoverview}

\begin{figure}[!t]
\vspace{0.2cm}
		\center
		\includegraphics[clip, trim=0cm 0cm 1cm 2cm, width=\columnwidth]{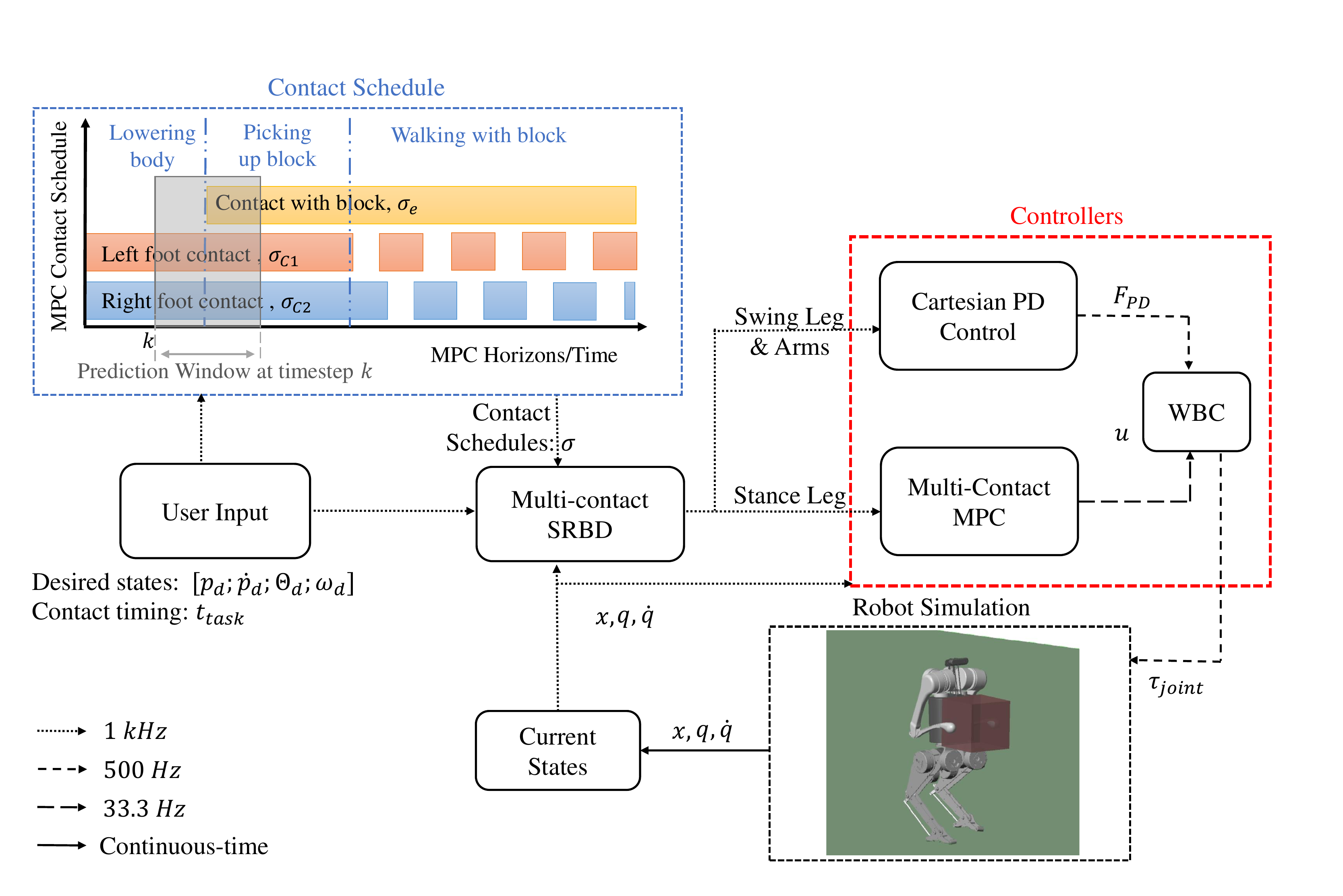}
		\caption{{\bfseries System Block Diagram} Control system architecture.}
		\label{fig:controlArchi}
		\vspace{-0.2cm}
\end{figure}

This section presents an overview of the control architecture of our proposed work, shown in Figure. \ref{fig:controlArchi}. 
In our approach, we leverage the contacting schedules and MPC prediction horizons to include smooth transitions between contact modes in loco-manipulation tasks. 

The user defines the contact timings of the loco-manipulation tasks along with desired states. 
The contact modes are represented by contact schedules. The contact mode/schedule information is very powerful due to the predictive nature of MPC that optimizes the current control input by the knowledge of future dynamics changes.  A visualization of the contact schedule based on the contact timing inputs in MPC is given in Figure. \ref{fig:controlArchi}. This contact schedule describes a task commanding the robot to pick up and walk with an object.
The time-varying multi-contact SRBD adapts based on the contact mode. The stance legs are controlled by MPC, while the swing legs and arms are controlled by the Cartesian PD controller. The optimal control inputs $\bm u \in \mathbb R^{10}$ are in terms of contact ground reaction forces $\bm F_n$ and moments $\bm M_n$ from multi-contact MPC.  Both $\bm F_{PD}\in \mathbb R^{12}$ and $\bm u$ are input to a Whole-body Control (WBC) to calculate the optimal joint torques $\bm \tau \in \mathbb R^{16}$ of the humanoid robot. 

The robot state feedback $\bm x$ include body Euler angles (roll, pitch, and yaw) ${\bm \Theta = [\phi,\:\theta,\:\psi]}^\intercal$, position $\bm p_c \in \mathbb R^{3}$, velocity of body CoM  $\dot{\bm p}_c \in \mathbb R^{3}$, and angular velocity $\bm \omega \in \mathbb R^{3}$. Joint feedback  includes the joint positions $\bm q \in \mathbb R^{16}$ and velocities $\dot{\bm q} \in \mathbb R^{16}$ of the humanoid robot.

\section{Proposed Approach}
\label{sec:dynamicsModel}

In this section, we investigate an optimal approach to represent the humanoid and object dynamics in an SRBD for multi-contact MPC. We also present the details of the multi-contact MPC framework.

\subsection{Dynamics Model}
\label{subsec:CSMPC}
Modifying the SRBD in force-based control has shown success in our previous works on bipedal robots \cite{li2021force}. In this work, we further investigate \rev{a suitable dynamics} model for the humanoid robot carrying a heavy object while considering the contact schedule for transition between different contact modes. 

\begin{figure}[t]
\vspace{0.2cm}
     \centering
     \begin{subfigure}[b]{0.15\textwidth}
         \centering
         \includegraphics[clip, trim=0cm 1cm 22.5cm 2cm, width=\columnwidth]{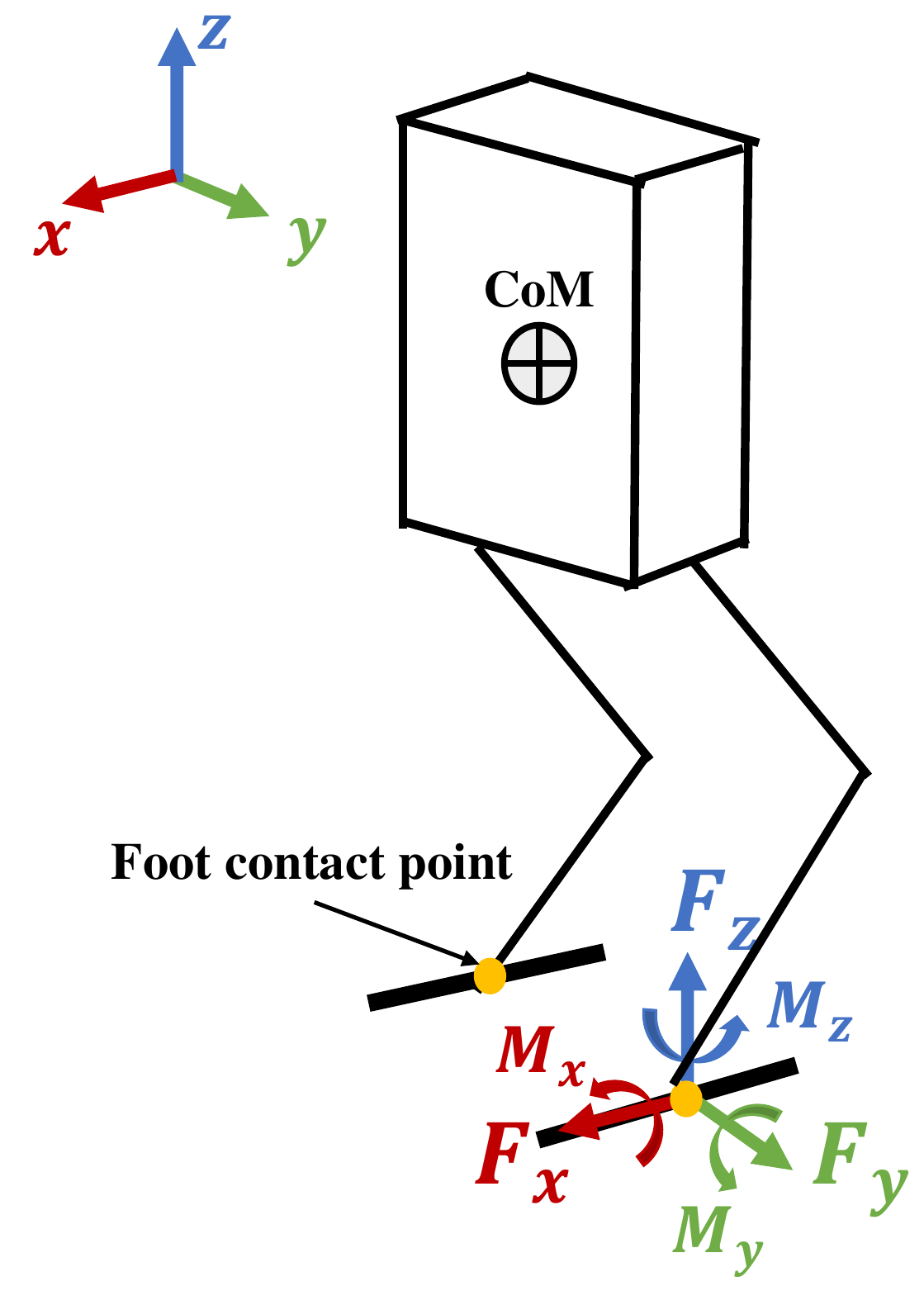}
         \caption{Bipedal SRBD \\ (Reference Baseline)}
         \label{fig:model1}
     \end{subfigure}
     \hfill
     \begin{subfigure}[b]{0.15\textwidth}
         \centering
         \includegraphics[clip, trim=23.3cm 1cm 0cm 2cm, width=\columnwidth]{Figures/dynamicModels.pdf}
         \caption{Humanoid SRBD 1 \\ (Combined Rigid Body)}
         \label{fig:model2}
     \end{subfigure}
     \hfill
     \begin{subfigure}[b]{0.155\textwidth}
         \centering
         \includegraphics[clip, trim=11.5cm 1cm 10.5cm 2cm, width=\columnwidth]{Figures/dynamicModels.pdf}
         \caption{Humanoid SRBD 2 \\ (External Force Model)}
         \label{fig:model3}
     \end{subfigure}
        \caption{{\bfseries SRBDs.} a) Baseline Bipedal SRBD in \cite{li2021force}  b) Combined rigid body dynamics  c) SRBD treating objects as external forces. Humanoid SRBD 2 is used in our proposed approach.}
        \label{fig:simplifiedDynamics}
        \vspace{-0.4cm}
\end{figure}

We first propose two different SRBD models for considering the object dynamics in humanoid robot SRBD. The baseline model that we developed our new models from is an SRBD for bipedal robot locomotion in \cite{li2021force}. Shown in Figure. \ref{fig:model1}, This SRBD treats the robot's upper body and hips as a rigid body to estimate the centroidal dynamics of this rigid body \cite{orin2013centroidal}. The ground reaction forces and moments at the foot locations are applied to the float base, 
\rev{$ \bm u=[\bm F_1;\:\bm F_2;\:\bm M_1;\:\bm M_2]$, where $ \bm F_n = [ F_{nx};\: F_{ny};\: F_{nz}], \bm M_n = [ M_{ny};\: M_{nz}], $ leg $n = 1, 2 $.} It is assumed that the robot has lightweight and low-inertia legs \cite{di2018dynamic,li2021force}. 

We form the following models with the assumption that the object's location and physical properties are known.

\subsubsection{\textbf{Model 1}}
 In the newly proposed SRBD for humanoid robot carrying an object in Figure. \ref{fig:model2}, we combine the upper body (rigid body formed by trunk, shoulders, and hips) with the object and treat them as a single combined rigid body, hence naming it the Combined Rigid Body model. The combined center of mass (CoM) location and rotational inertia are updated in real-time base on the location of the object $\bm p_o$ and upper body $\bm p_{c}$. We use the parallel axis theorem \cite{hass} to calculate the rotational inertia $\bm I_b$ of the rigid body. The combined CoM location $\bm p_h \in \mathbb R^{3}$ is
\begin{align}
\bm p_h = \frac{\bm p_{c} m_{ub} + \bm p_{o} m_{o}}{ m_{ub} +  m_{o}},
\end{align}
where $m_{ub}$ and $m_{o}$ are the upper body mass and object mass. 

In this proposed configuration, the combined rigid body CoM location is in between trunk CoM and Object CoM locations. Hence, to align the CoM location with the hip and foot location vertically for better balancing performance, the upper body pitches backward. This also mimics how a human carries a heavy box and remains balanced. 

\subsubsection{\textbf{Model 2}}
In the second approach we investigated, naming it External Force Model, we simplified the object to an external force $\bm F_{ext} \in \mathbb R^{3}$ applied to the rigid body, shown in Figure. \ref{fig:model3}. This external force is the only added terms in the formulation of SRBD, hence reducing the number of variables that has sudden change in manipulation tasks. 

\gpt{Model 1 has a drawback, which is that when an object is added to the upper body during manipulation, the SRBD formulation explicitly expresses sudden nonlinear changes in state variables, particularly in the combined CoM location, pitch angle, and physical properties of the rigid body. As a result, the MPC problem faces challenges in finding optimal solutions due to these sudden, significant, and nonlinear changes.}
Whereas Model 2 tackles this problem by only having one added term (i.e. external force) to the dynamics for simplicity and linearity when dynamics change between contact modes. 
Comparisons of the two investigated approaches are shown in Section. \ref{sec:Results}. 

\begin{remark}
\vspace{0.1cm}
\textit{The humanoid SRBD we decided to use for the rest of the work is Model 2: External Force Model.}
\vspace{0.1cm}
\end{remark}

 Developed from the \cite{nguyen2019optimized} and \cite{li2021force} The multi-contact SRBD is expressed as follows,
\begin{multline}
\label{eq:simplifiedDynamics}
\left[\begin{array}{cccc} 
\sigma _{C1}(t)\mathbf {I}_{3}  & \sigma _{C2}(t) \mathbf {I}_{3} &  \mathbf {0}_{3\times2} &  \mathbf {0}_{3\times2}   \\ 
\sigma _{C1}(t) \bm r_1\times  & \sigma _{C2}(t) \bm r_2\times & 
\sigma _{C1}(t)  \mathbf L & \sigma _{C2}(t) \mathbf L \end{array}  \right] \bm  u \\
+ \sigma_{e}(t)\left[\begin{array}{c} \bm F_{ext}(t) \\ \bm r_e\times \bm F_{ext}(t) \end{array}  \right]= 
 \left[\begin{array}{c} m (\ddot{\pos}_{c} +\bm{g}) \\ \frac{d}{dt}(\bm I_G  {{\bm \omega}}) \end{array} \right],
\end{multline}
where $\bm r_i\times$ and $\bm r_e\times$ stands for the skew-symmetric matrix of distance vector from upper body CoM $\bm p_c$ to foot $\bm p_i$ and to external object CoM location, $\bm I_G$ is the rotational inertia of the upper body expressed in the world frame. $\mathbf L$ is moment selection matrix, $\mathbf L = [0, 0; 1, 0; 0, 1]$.

The time-dependent contact-schedule terms $\sigma(t)$ for each task are expressed in either 1s or 0s to reflect the task schedule. Different combinations of these contact schedules reflect different contact modes in the dynamics model.




\subsection{Multi-contact MPC}
\label{subsec:CSMPC_form}

Now we present the multi-contact MPC formulation that employs the SRBD proposed in Section.\ref{subsec:CSMPC}. We choose the state variables as $[{\bm \Theta};{\bm p}_c;{\bm \omega};\dot {{\bm p}}_c]$ and control inputs as $\bm U = [\bm u; \bm F_{ext}]$, then the simplified dynamics equation can be represented as
\begin{align}
\label{eq:simpDyn}
\frac{d}{dt}\left[\begin{array}{c} {\bm \Theta}\\{\bm p}_c\\{\bm \omega}\\\dot {{\bm p}}_c \end{array} \right] 
= \bm A \left[\begin{array}{c} {\bm \Theta}\\{\bm p}_c\\{\bm \omega}\\\dot {{\bm p}}_c \end{array} \right] + \bm B \bm U + \left[\begin{array}{c} \mathbf 0_{3\times1}\\\mathbf 0_{3\times1}\\\mathbf 0_{3\times1}\\\bm g \end{array} \right] 
\end{align}

\begin{align}
\setlength\arraycolsep{2pt}
\label{eq:A}
\bm A = \left[\begin{array}{cccc} 
\mathbf 0_3 & \mathbf 0_3 & \mathbf R_b & \mathbf 0_3 \\
\mathbf 0_3 & \mathbf 0_3 & \mathbf 0_3 & \mathbf I_3 \\
\mathbf 0_3 & \mathbf 0_3 & \mathbf 0_3 & \mathbf 0_3 \\
\mathbf 0_3 & \mathbf 0_3 & \mathbf 0_3 & \mathbf 0_3  \end{array} \right], 
\mathbf R_b = \left[\begin{array}{ccc}
{c_\theta}{c_\psi} & -{s_\psi} & 0 \\
{c_\theta}{s_\psi} & c_\psi & 0 \\
-s_\theta & 0 & 1  \end{array} \right] 
\end{align}

\begin{align}
\label{eq:B}
\bm B = \left[\begin{array}{ccccc} 
\mathbf 0_3 & \mathbf 0_3 & \mathbf 0_{3\times2} & \mathbf 0_{3\times2} & \mathbf 0_3\\
\mathbf 0_3 & \mathbf 0_3 & \mathbf 0_{3\times2} & \mathbf 0_{3\times2}  & \mathbf 0_3 \\ 
\frac{\sigma _{C1} \bm r_1\times}{\bm I_G}  & \frac{\sigma _{C2} \bm r_2\times}{\bm I_G} & 
\frac{\sigma _{C1}\mathbf L}{\bm I_G}   & \frac{\sigma _{C2} \mathbf L}{\bm I_G} & \frac{\sigma _{e}\bm r_e\times}{\bm I_G}\\
\frac{\sigma _{C1}\mathbf {I}_{3}}{m_{ub}}  & \frac{\sigma _{C2} \mathbf {I}_{3}}{m_{ub}} &  \mathbf {0}_{3\times2} &  \mathbf {0}_{3\times2} & \frac{\sigma _{e}\mathbf {I}_{3}}{m_{ub}} \end{array}  \right]
\end{align}
where $s$ denotes sine operator, and $c$ denotes cosine operator. Note that $\mathbf R_b$ is not invertible at $\theta = \pm90^\circ$. We intend to not allow the robot to have a pitch angle of $\pm90^\circ$ in any tasks.

To guarantee the linear relation in the formulation of Equation (\ref{eq:simpDyn}), we choose to include gravity $\bm g \in \mathbb{R}^3$ as a dummy variable in decision variables $\bm x = [{\bm \Theta};{\bm p}_c;{\bm \omega};\dot {{\bm p}}_c; \bm g]\in \mathbb{R}^{15}$. This ensures the presence of gravity and forms a state-space equation with continuous-time matrices ${\hat{\bm A_c}}$ and ${\hat{\bm B_c}}$: 
\begin{figure*}[!t]
	\vspace{0.2cm}
    \center
    \includegraphics[width=.8\textwidth]{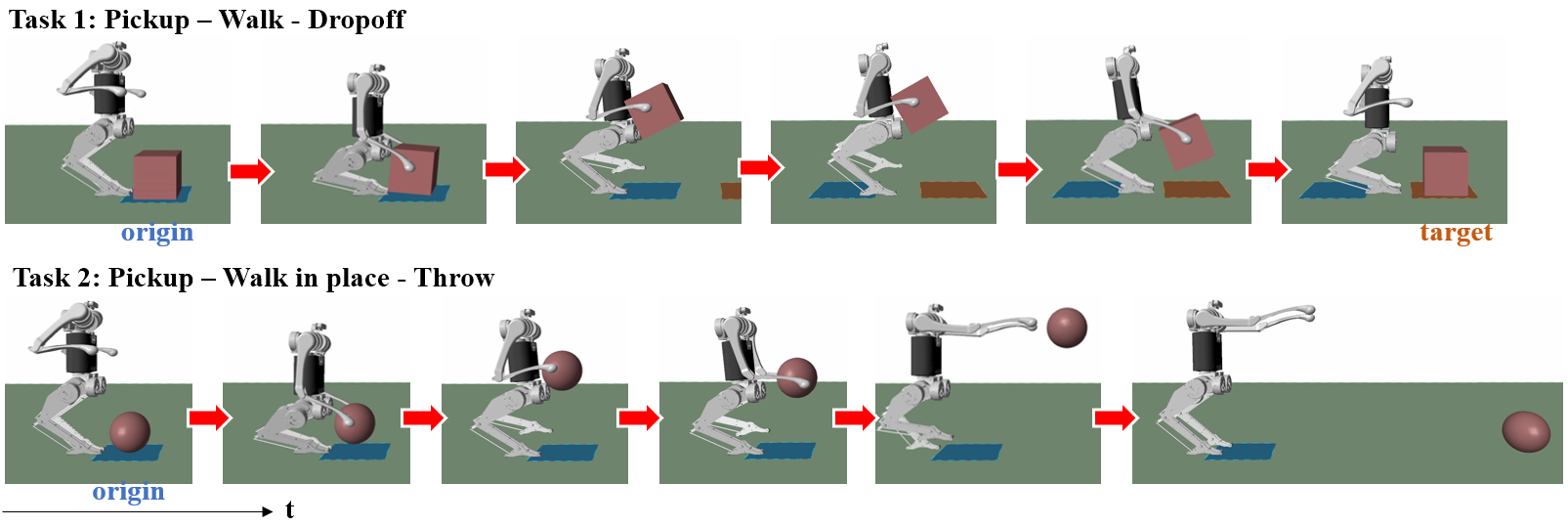}
    \caption{{\bfseries Simulation Snapshots}  1) Pickup, walk, and drop off a 4 $\unit{kg}$ box object, 2) Pickup, walk in-place, and throw a 2 $\unit{kg}$ spherical object. The Blue region represents the origin of the object. The Orange region represents the target local for drop-off. }
    \label{fig:snapshots}
    \vspace{-0.7cm}
\end{figure*}
\begin{align}
\label{eq:linearSS}
\dot{{\bm { x}}}(t) =  {\hat{\bm A_c}} {{\bm {x}}} +  {\hat{\bm B_c}} \bm U.
\end{align}

 The formulation of the MPC problem with finite horizon $k$ is written as follows. The objective of the problem is to drive state $\bm x$ close to the desired input and minimize $\bm U$, weighted by matrices $\bm Q_i \in \mathbb R^{15\times 15}$ and $\bm R_i \in \mathbb R^{10\times 10}$.
\begin{align}
\label{eq:MPCform}
\underset{\bm{x,U}}{\operatorname{min}}   \:\:  & \sum_{i = 0}^{k-1}(\bm x_{i+1}-  \bm x_{i+1}^{ref})^T\bm Q_i(\bm x_{i+1}- \bm x_{i+1}^{ref}) + \bm{R}_i\| \bm{U}_i \|
\end{align}
\begin{subequations}
\setlength\abovedisplayskip{-5pt}
\begin{align}
\label{eq:dynamicCons}
\:\:\operatorname{s.t.} \quad  {\bm {x}}[i+1] = \bm {\hat{A}}[i]\bm x[i] + \bm {\hat{B}}[i]\bm U[i], \\
\label{eq:frictionCons}
\nonumber 
-\mu  {F}_{nz} \leq  F_{nx} \leq \mu {F}_{nz} \quad \quad\\
-\mu {F}_{nz} \leq  F_{ny} \leq \mu {F}_{nz} \quad \quad\\
\label{eq:forceCons}
0<  {F}_{min} \leq  F_{nz} \leq  {F}_{max} \quad \quad\\
\label{eq:forceCons2}
\bm {F}_{ext} =  \left[\begin{array}{ccc} 0 & 0 & m_og \end{array} \right]^\intercal \quad \quad
\end{align}
\end{subequations}

Equation (\ref{eq:dynamicCons}) to (\ref{eq:forceCons}) are constraints of the MPC problem. Equation (\ref{eq:dynamicCons}) is an equality constraint of the linearized dynamics equation in discrete-time at $i$th time-step derived from equation (\ref{eq:linearSS}). Equation (\ref{eq:frictionCons}) describes inequality constraints on the contact friction pyramid. Equation (\ref{eq:forceCons}) describes the bounds of reaction forces. Equation (\ref{eq:forceCons2}) is an equality constraint that ensures the external force from the control input is equal to the gravitational force of the object.

The translation of the proposed MPC problem into Quadratic Programming (QP) form to be efficiently solved \rev{can be found in \cite{jerez2011condensed}.}

\subsection{Low-level Control}
\label{subsec:lowlevel}

While the multi-contact MPC primarily provides optimal ground reaction force-and-moments to the stance legs during locomotion, we choose to use Cartesian-space PD control to manipulate arms and swing foot $n$ based on the tasks and contact modes. And then map all controller forces to joint torques by WBC.

The desired swing foot location $\bm p_{f,n}^{des} \in \mathbb R^{3}$ follows a heuristic foot placement policy based on spring-loaded inverted pendulum \cite{raibert1986legged}, with addition of the consideration of desired CoM linear speed, $\dot{\bm p}_c^{des}$ \cite{kim2019highly},
\begin{align}
\label{eq:footPlacement}
\bm p_{f,n}^{des} =  \bm p_{c} + \dot{\bm p}_c \Delta t/2 + k(\dot{\bm p}_c-\dot{\bm p}_c^{des}),
\end{align}
where $\Delta t$ is the gait period, and $k$ is a scaling factor for tracking desired linear velocity. (e.g., push-recovery)

The $n$th swing foot force by Cartesian PD control is 
\begin{align}
\label{eq:pdlaw}
\bm F_{swing,n}=\bm K_P(\bm p_{f,n}^{des}-\bm p_{f,n})+\bm K_D(\dot{\bm p}_{f,n}^{des}-\dot{\bm p}_{f,n})
\end{align}

Similarly, to control the $m$th hand of the robot to move to desired task locations $\bm p_{h,m}^{des} \in \mathbb R^{3}$, the hand force is 
\begin{align}
\label{eq:pdlaw2}
\bm F_{hand,m}=\bm K_P(\bm p_{h,1}^{des}-\bm p_{h,m})+\bm K_D(\dot{\bm p}_{h,m}^{des}-\dot{\bm p}_{h,m})
\end{align}

Using WBC to map optimal ground reaction forces to joint torques on legged robots is an established approach in many related works (e.g.,\cite{kim2019highly, chignoli2021humanoid}). We use WBC as a low-level controller to synchronize the hybrid controller inputs described from MPC and Cartesian PD controllers.

We extend the second humanoid SRBD (external force model) in Section. \ref{subsec:CSMPC} to the whole-body dynamics of the humanoid robot with the object. Similarly, by simplifying the object dynamics as an external force applied to the robot, the full joint space equation of motion is,
\begin{align}
\label{eq:EOM1}
\mathbf M \ddot{\mathbf q} + \mathbf C + \mathbf g = \left[\begin{array}{c}  \mathbf 0 \\  \bm \tau \end{array} \right] 
+ \bm \tau_f
\end{align}
\begin{multline}
\label{eq:EOM2}
\bm \tau_f =  \bm J_c^\intercal \left[\begin{array}{c}  \sigma _{C1}(t)\bm F_1 \\  \sigma _{C2}(t)\bm F_2 \\ \sigma _{C1}(t)\bm M_1 \\ \sigma _{C2}(t)\bm M_2 \end{array} \right] 
 + \bm J_{PD}^\intercal \left[\begin{array}{c}  \sigma _{H1}(t)\bm F_{hand,1} \\ \sigma _{H2}(t)\bm F_{hand,2} \\  \sigma _{C2}(t)\bm F_{swing,1} \\ \sigma _{C1}(t)\bm F_{swing,2} \end{array} \right] \\
+ \sigma _{e}(t) \bm J_{e}^\intercal \bm F_{ext}(t)
\end{multline}
where $\mathbf M \in \mathbb R^{22\times 22}$ is the generalized mass matrix, $\mathbf C$, $\mathbf g \in \mathbb R^{22}$ are the Coriolis and gravity forces. $\ddot{\mathbf q}$ is a vector space containing acceleration of floating base trunk $\ddot{\mathbf q_b}\in \mathbb{R}^{6}$ and joints $\ddot{\mathbf q_j}\in \mathbb{R}^{16}$, as described in \cite{kim2019highly}. $\bm J_c$, $\bm J_{PD}$, and $\bm J_{e}$ are the Jacobians describing the foot contacts, Cartesian PD control end-effector locations (i.e. hands are feet), and external force location (i.e. CoM of the object). 

The WBC can be simplified and formed into a QP problem for efficient execution and high frequency. The desired command in WBC is based on the desired states of the robot $\bm x_c^{des} = [\bm p_c^{des}; \bm \Theta^{des}]$, by a simple PD control law:
\begin{align}
\label{eq:desAcc}
\ddot{\bm x}_c^{des} = \bm K_p^{WBC}(\bm x_c^{des} - \bm x_c) + \bm K_d^{WBC}(\dot{\bm x}_c^{des} - \dot{\bm x}_c)
\end{align}

The desired acceleration command $\bm \ddot{x}_c^{des}$ is translated into task-space $\ddot {\mathbf {q}}_{cmd}$ by an inverse-kinematic-based null space projection technique described in \cite{kim2019highly}. 

Now the WBC-QP problem to compute the minimized relaxation components of MPC ground reaction force $\Delta \bm u$ and joint acceleration command $\Delta \ddot{\mathbf q}$ is as follows,
\begin{align}
\label{eq:WBC-QP}
\underset{{\Delta \ddot{\mathbf q},\Delta \bm u}}{\operatorname{min}}   \:\:  & 
\Delta \ddot{\mathbf q}^\intercal {\mathbf H} \Delta \ddot{\mathbf q} + \Delta \bm u^\intercal {\mathbf K} \Delta \bm u
\vspace{0.5cm}
\end{align}
\begin{subequations}
\setlength\abovedisplayskip{-6pt}
\begin{align}
\label{eq:WBC_cons1}
\nonumber
\operatorname{s.t.} \quad 
\mathbf S_{b}\{\mathbf M (\Delta \ddot{\mathbf q} + \ddot{\mathbf q}_{cmd}) + \mathbf C + \mathbf g \\ 
- \bm \tau_f(t, \Delta \bm u , \bm u)\} = \mathbf 0 \\
\label{eq:WBC_cons3}
\quad  \quad  \quad  \bm u_{min} \leq \Delta \bm u + \bm u \leq \bm u_{max} \quad\\
\label{eq:WBC_cons4}
\quad  \quad  \quad  \bm \tau_{min} \leq \bm \tau \leq \bm \tau_{max} \quad
\end{align}
\end{subequations}

In equation (\ref{eq:WBC-QP}), $\mathbf H \in  \mathbb{R}^{16\times16}$ and $\mathbf K \in  \mathbb{R}^{10\times10}$ are diagonal weighting matrices for each objective. Equation (\ref{eq:WBC_cons1}) is a dynamics constraint formed by equation \ref{eq:EOM1} in order to control the floating base dynamics. Selection matrix $\mathbf S_{b}$ consists of 1s and 0s to identify the float base joints.

Finally, the optimal joint torque $\bm \tau$ can be calculated as
\begin{align}
\label{eq:torque}
\left[\begin{array}{c}  \bm 0 \\  \bm \tau \end{array} \right] 
= \mathbf M (\Delta \ddot{\mathbf q} + \ddot{\mathbf q}_{cmd}) + \mathbf C + \mathbf g - \bm \tau_f(t, \Delta \bm u , \bm u)
\end{align}

where $\bm \tau_f$ is a time-dependent term that summarizes all external forces applied to the system based on the contact schedules, shown in equation \ref{eq:EOM2}. The swing foot and hand forces are feed-forward from the Cartesian PD controller, and optimal reaction force-and-moments are $\Delta \bm u+\bm u$.

\section{Results}
\label{sec:Results}

In this section, we will present highlighted results for validation of our proposed humanoid dynamics model with object dynamics and the multi-contact MPC framework in humanoid object-carrying simulations. The reader is encouraged to watch the supplementary simulation videos\footnote{\url{https://youtu.be/V8PIpE2YGhw}}.

We validate our proposed approach in a high-fidelity physical-realistic simulation framework in MATLAB Simulink with Simscape Multibody library. We also use Spatial v2 software package \cite{featherstone2014rigid} to acquire coefficient matrices of dynamics equations for WBC. In this simulation, we assume the state information and physical properties of the object are known.

\rev{The weighting matrices in MPC and PD gains in WBC are universal in the following simulation results, where}
\begin{flalign*}
\nonumber
\bm Q_i = \text{diag}[1500\:2000\:1000\:1000\:1000\:1000\:1\:3\:10\:1\:1\: 1 \:1\:1\:1],\: \\
\nonumber
\bm R_i = \text{diag}[1\:1\: 1 \:1\:1\: 1 \:5\:5\: 5 \:5\:5\: 5 \:]\times10^{-4} \quad \quad \quad  \quad \quad \quad \quad \quad \: \: \: \\
\nonumber
\bm K_p^{WBC} = \text{diag}[200\:200\: 500 \:1000\:1500\: 1000] \quad \quad \quad \quad \quad \quad \: \: \: \: \: \\
\nonumber
\bm K_d^{WBC} = \text{diag}[20\:20\: 30 \:30\:30\: 30] \quad \quad \quad \quad \quad \quad  \quad \quad  \quad \quad \quad \: \: \:
\end{flalign*}

\begin{figure}[!t]
\vspace{0.2cm}
		\center
		\includegraphics[clip, trim=1.8cm 4.7cm 8.4cm 2cm, width=0.85\columnwidth]{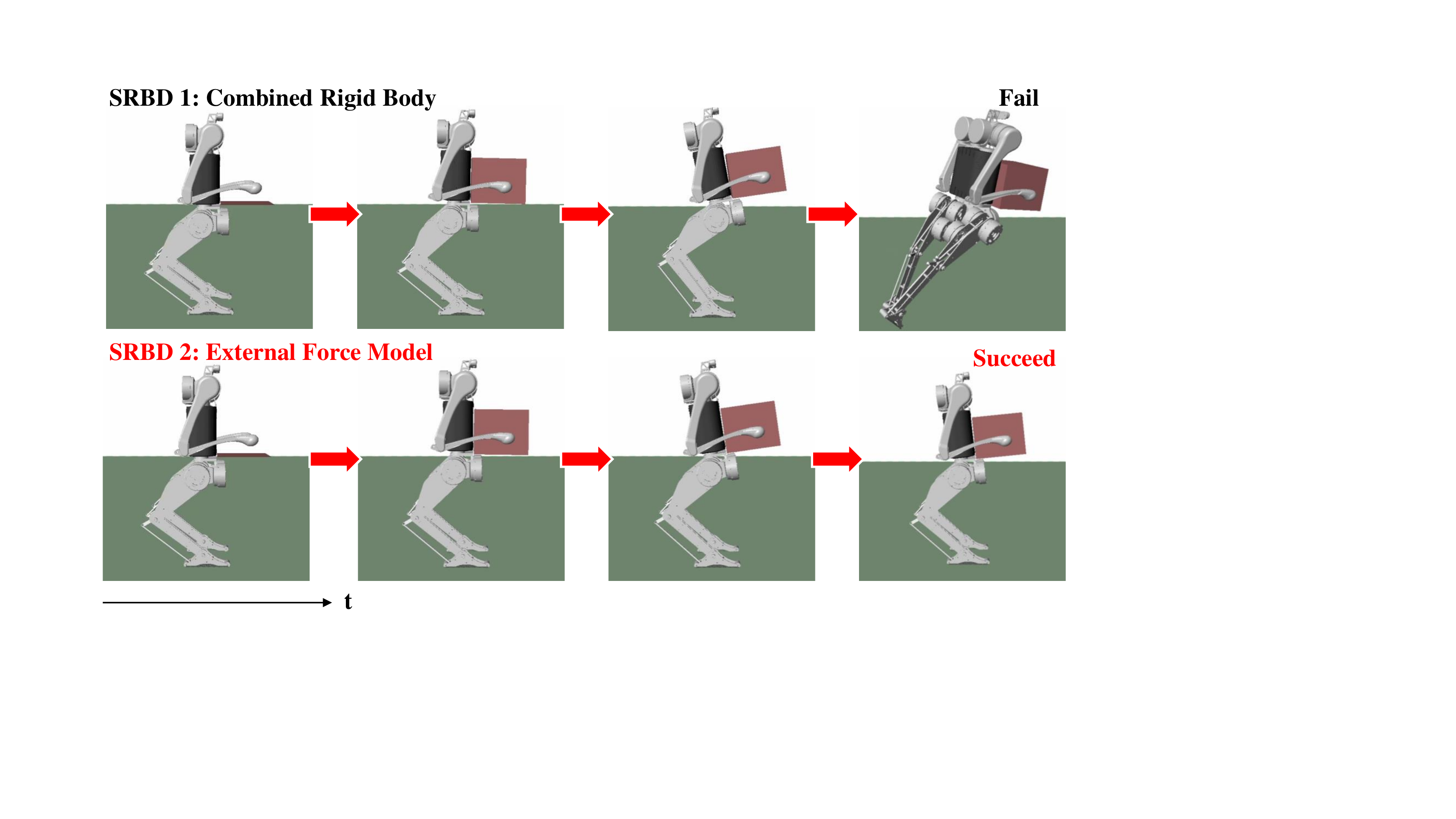}
		\caption{{\bfseries Comparison of Dynamics Models} Simulation snapshots of 1) Hybrid Rigid Body Model vs. 2) External Force Model in humanoid balancing with 5 $\unit{kg}$ weight block. }
		\label{fig:comparison}
		\vspace{-.4cm}
\end{figure}

Firstly, we present the comparison between the dynamics models we investigated in Section. \ref{sec:dynamicsModel} in a simple simulation of a humanoid robot balancing while holding a 5 $\unit{kg}$ box. We compare the approaches of using the combined rigid body model in Figure. \ref{fig:model2} and external force model in Figure. \ref{fig:model3}. Figure. \ref{fig:comparison} shows the comparison simulation snapshots, with the combined rigid body model in controllers, the robot is not able to recover from leaning forward when the external weight is applied. However, controllers using the external force model handle the applied external weight well and are able to adapt the robot to a favorable pose to carry the weight. (i.e. leaning back) With this framework, our humanoid robot can carry up to 14 $\unit{kg}$ ($82\%$ robot mass) while standing still. Figure. \ref{fig:torque} shows the joint torque plots of this simulation.

\begin{figure}[!t]
\vspace{0.2cm}
		\center
		\includegraphics[clip, trim=0.2cm 4cm 6.2cm 4cm, width=0.85\columnwidth]{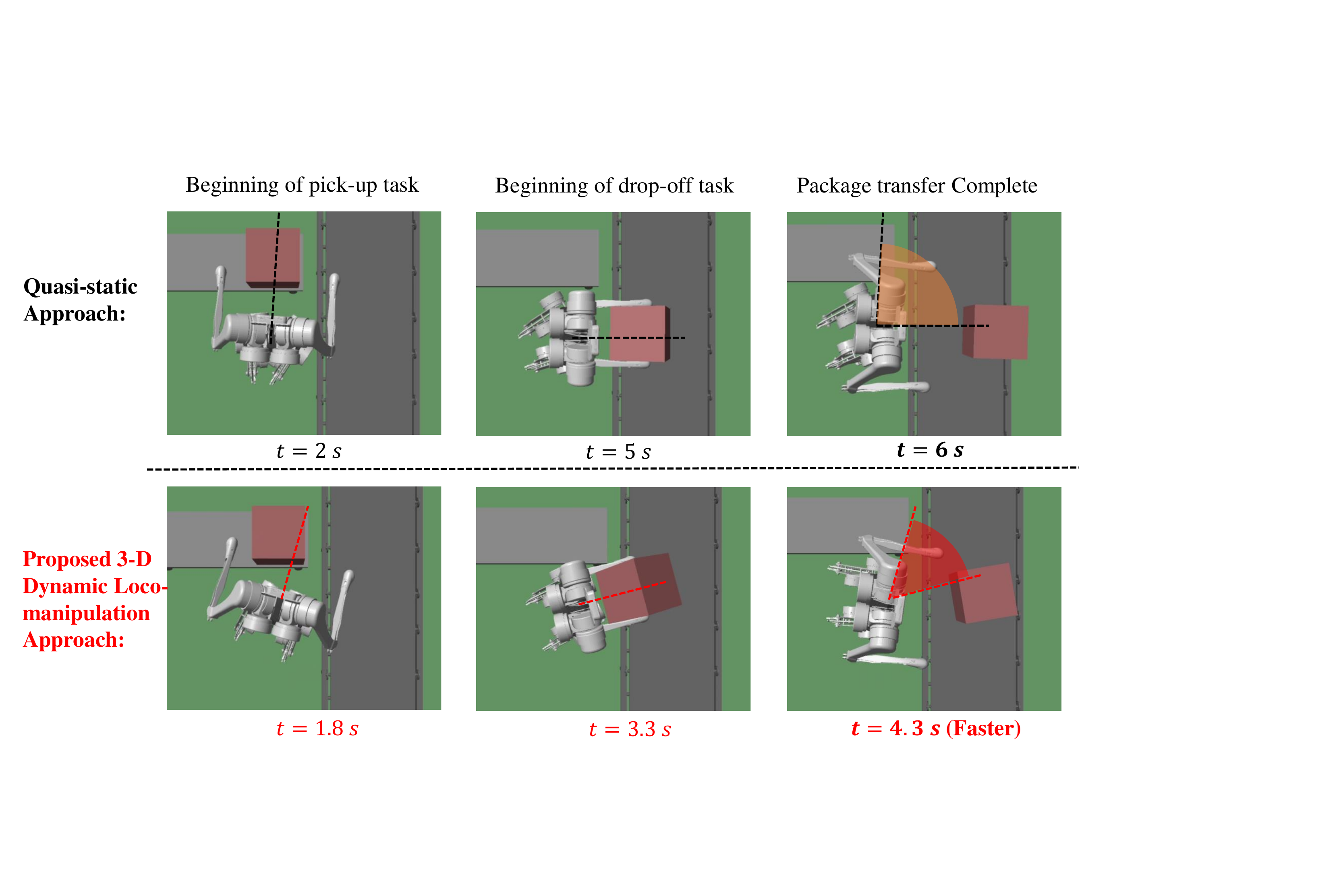}
		\caption{{\bfseries Package Transfer Simulation Snapshots} Comparisons of quasi-static approach that separates locomotion and manipulation, and proposed multi-contact MPC loco-manipulation approach}
		\label{fig:90turn}
		\vspace{-.4cm}
\end{figure}

Next, we present 2-D loco-manipulation examples in simulation following our proposed multi-contact MPC framework. Shown in Figure. \ref{fig:snapshots}, the top snapshots represent the simulation when giving contact schedule information of tasks following the sequence of pick up, walk, and drop off a 4 $\unit{kg}$ box-shaped object. The contact schedule information is given to the controllers offline based on the target location and command tasks by the user. The bottom snapshots in Figure. \ref{fig:snapshots} represents a simulation of picking up, walking in-place, and throwing a 2 $\unit{kg}$ spherical object. It can be observed that our proposed framework provides good performance when multiple tasks take place concurrently. The framework synchronizes and bridges contact modes by long prediction horizons and high frequency desired state tracking. In the above simulations, the MPC samples at 0.03 $\unit{s}$ and has a prediction horizon $k = 20$. 

We would also like to demonstrate the advantages of our multi-contact MPC framework on humanoid loco-manipulation in logistic applications. The example task is to transfer a package from a table to a conveyor belt, following:
\begin{enumerate}
    \item Turn $90^\circ$ counterclockwise 
    \item Pick up a package on the table
    \item Turn $90^\circ$ clockwise 
    \item Drop off the package on the conveyor belt
\end{enumerate}

The traditional approaches on many existing humanoids are to follow the task sequence and stand still while manipulating the package quasi-statically (e.g.,\cite{digityoutube} \cite{atlasyoutube}). With the proposed approach, we can allow the robot to combine 3-D locomotion (for turning) with dynamic manipulation to complete such task with a considerably faster speed. Snapshots of the simulation top view are presented in Figure. \ref{fig:90turn}. The dotted lines in this figure represent the direction the robot is facing at the beginning of each task, and the shaded sectors represent the yaw angle differences of picking-up and dropping-off starting timings. 

\gpt{Due to the absence of synchronized contact schedules and transitions between contact modes, the quasi-static approach involves commanding the robot to stand still while picking up or dropping off packages, and turning in-place with locomotion.}
On the other hand, in a multi-contact MPC where all contact modes and transitions are known in the prediction horizons, we can allow 3-D loco-manipulation with the object more dynamically. This allows earlier manipulation task timings because the robot does not need to stop during turning yet performs more efficiently than the quasi-static approach. It is observed that the yaw angle difference between picking-up and dropping-off task timings can be up to 30$^\circ$ less than that in the quasi-static approach. The proposed approach is also 1.7 $\unit{s}$ faster in this proposed example. Figure. \ref{fig:torque2} shows the left leg joint torques in this example with the proposed approach. All torque values are in joint torque limits in such dynamic motion.



\begin{figure}[!t]
	\vspace{0.2cm}
     \center
     \begin{subfigure}[b]{0.4\textwidth}
         \center
		\includegraphics[clip, trim=3cm 10.6cm 3cm 10.6cm, width=\columnwidth]{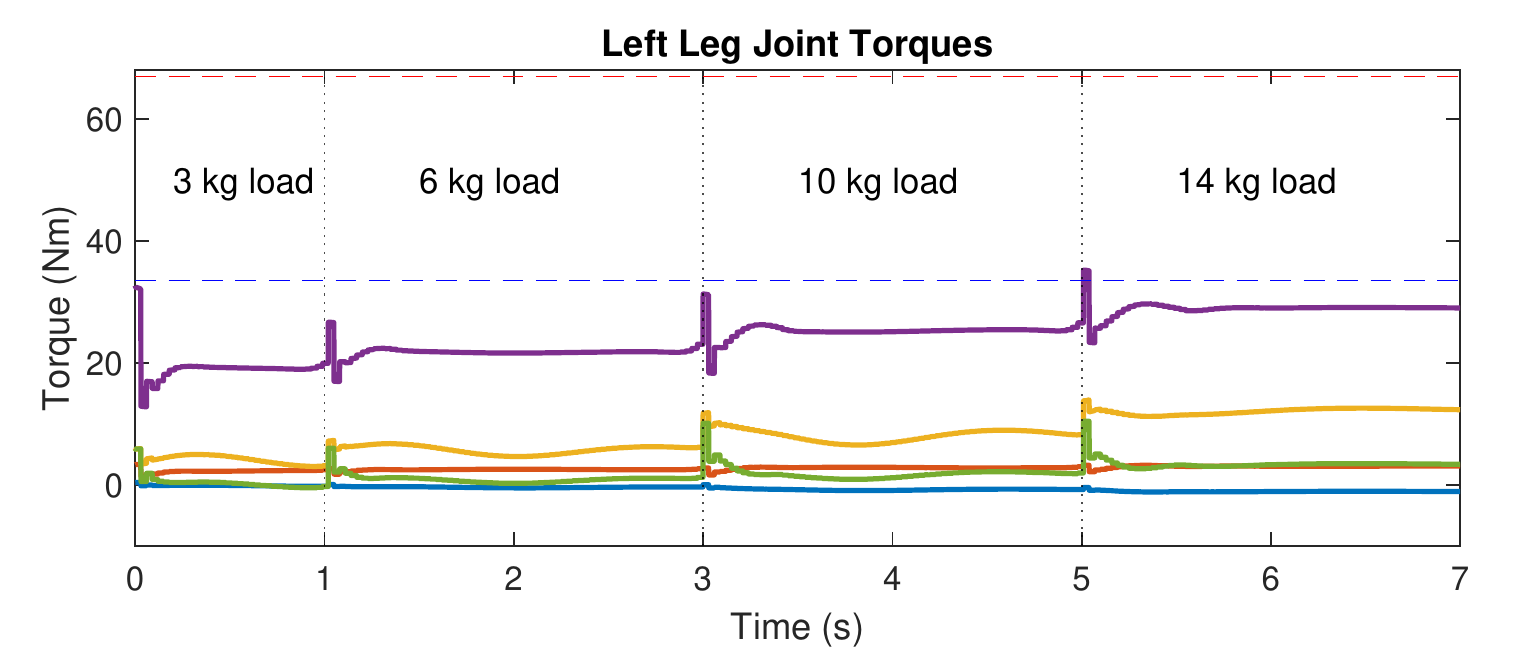}
		\vspace{-0.7cm}
		\caption{ Simulation of Load-carrying up to 14 $\unit{kg}$}
		\label{fig:torque}
		\vspace{0.2cm}
     \end{subfigure}
     \vspace{-0.2cm}
     \\
     \begin{subfigure}[b]{0.4\textwidth}
         \center
		\includegraphics[clip, trim=3cm 9cm 3cm 10cm, width=\columnwidth]{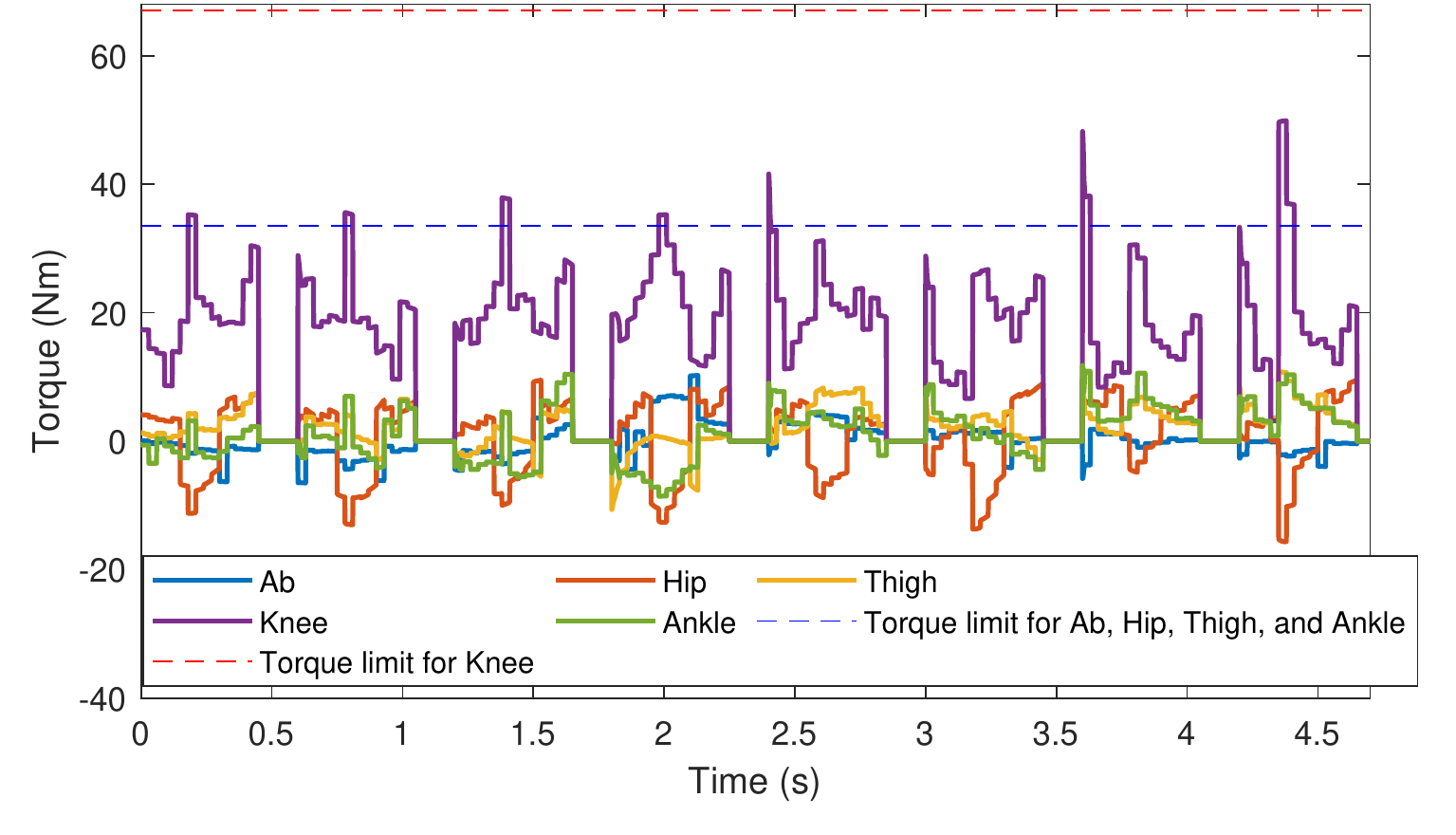}
		\vspace{-0.7cm}
		\caption{Simulation of proposed approach in Figure. \ref{fig:90turn}}
		\label{fig:torque2}
     \end{subfigure}
        \caption{{\bfseries Torque Plots of Left Leg Joints}}
        \label{fig:torques}
        \vspace{-0.5cm}
\end{figure}


\section{Conclusions}
\label{sec:Conclusion}

In conclusion, we introduced a multi-contact MPC framework to tackle the problem of contact mode transitions in humanoid dynamic loco-manipulations.  We investigated the most optimal dynamics model for humanoid loco-manipulation when considering object dynamics. By treating the object as an external force in dynamics, we eliminated sudden and significant formulation changes in MPC formulation. With the proposed method, we allowed the humanoid robot to complete 3-D multi-task dynamic loco-manipulations. The proposed method is more efficient than  the quasi-static approach on humanoid robots in applications such as package transferring in logistics.

\section{Acknowledgment}
\label{sec:Acknowledgement}

The authors would like to thank Han Gong, Junchao Ma, and Manas Shah for contributing to this work.

\balance
\bibliographystyle{ieeetr}
\bibliography{reference.bib}

\end{document}